\documentclass[12pt]{article}

\usepackage{a4wide}
\usepackage{graphics}
\usepackage{color}
\usepackage{algorithm}
\usepackage{algorithmic}

\newcommand{\D}{I\!\!D}
\renewcommand{\P}{I\!\!P}
\newcommand{\R}{I\!\!R}
\newcommand{\W}{{\cal W}}
\newcommand{\G}{{\cal G}}
\renewcommand{\d}{{\cal D}}
\newcommand{\T}{{\cal T}}
\newcommand{\ccs}{{BPO's}}
\newcommand{\cc}{{BPO}}
\newcommand{\DEF}{\stackrel{\rm def}{=}}

\title{\bf A Metric to Classify {\em Style} of Spoken Speech}
\author{Sunil Kopparapu\thanks{Contact e-mail:
SunilKumar.Kopparapu@TataInfotech.Com}, Saurabh Bhatnagar, K. Sahana,
Sathyanarayana, \and 
Akhilesh Srivastava,  P.V.S. Rao\\[1ex]
Cognitive Systems Research Laboratory\\
Tata Infotech Limited, Navi Mumbai \\[1ex]
http://www.tatainfotech.com}
\date{}

\begin{document}
\maketitle

\begin{abstract} 
The ability to classify spoken speech based on the style
of speaking is an important problem. With the advent of \ccs\ in
recent times, specifically those that cater to a population other than the
local population, it has become necessary for \ccs\ to identify
people with certain style of speaking (American, British etc). Today 
\ccs\ employ accent analysts to identify people having the
required style of speaking. This process while involving human bias, it is
becoming increasingly infeasible because of the high attrition rate in the
\cc\ industry. In this paper, we propose a new metric, which
robustly and accurately helps classify spoken speech based on the style of
speaking. The role of the proposed metric is substantiated by using it to
classify real speech data collected from over seventy different people
working in a \cc. We compare the performance of the metric
against human experts who independently carried out the classification
process. Experimental results show that the performance of the system using the
novel metric performs better than two different human expert.
\end{abstract}

\section{Introduction}

\ccs\ (Business Process Outsourcing) centers are increasingly finding their way because of the
increased quality consciousness, particularly in the service industry
segment. Development in the area of telecommunications make it feasible
for the \ccs\ to be located in regions which it is servicing other
than the local population. In addition socio-economic reasons justify the
geographical location of \ccs\ anywhere without the people being
serviced being aware of it. This has led to a spate of \ccs\
cropping up in developing countries where there exists a large population
that can speak the language of the people not necessarily in the same
style. For this reason, there is no definite recruitment qualification
that one should possess to join a \cc, except that, one be able
to speak in the style of the population that the \cc\ services.
The increase in number of \ccs\ and the no specific
qualification requirement, leads to a situation of total influx, people
are always on the move (high attrition). This leads to the requirement of
a constant recruitment process at the \ccs. Today, \ccs\ with no exception, employ accent analyst to select candidates. The
accent analyst judges the suitability of a candidate by analyzing the
speaking style of the candidate. The process of recruitment is time
consuming (on an average only about 7\% of the candidates appearing for the
interview are selected) and is prone to human bias. There is a need for an
automatic system that can measure the candidates speaking style or more
precisely, classify the candidates speaking style as being suitable (good), trainable
(average) or unsuitable (bad).

Often one is able to make out the speakers background (American, British,
Indian etc) by just listening to the spoken speech of the person. In
addition, one is also able to tell if the person is speaking well or not,
even in the absence of knowledge of the language being spoken. Thus, it is
possible for a human to categorize speakers based of their speaking style
by listening to their speech. A trained human is able to perform this task
of classification better because he is aware of the nuances of what to be
on the lookout for which identifies a well spoken speech. An ideal
system\footnote{Essentially the system would be built by first analyzing
and deriving rules by listening to spoken speech samples. These rules
would enable development of the system to determine the quality of
speech.} would be the one that has the ability to classify people based on
their speaking style by looking at their free-spoken speech. While  work
is on at the Cognitive Systems Research Laboratory of Tata Infotech, the
development of such a system is still premature.

In this paper, we propose a system that can be used to classify people
based on their speaking style\footnote{To assist \cc\
recruitment process.}. The heart of the system is the use of a new metric,
which captures the speaking style of a person. Further, we describe
the construction and use of such a  metric. In this paper, we aim at
developing a system that is able to categorize the speaking style of a
person by analyzing predetermined set of words and sentences\footnote{The
speaker would be asked to speak a carefully selected list of words and
sentences, which would be used by the system to analyze the style of
speaking.}.

\section{Metric to Classify spoken speech}
\label{sec:metric}

The speaking style and articulatory capability of spoken speech can be 
assessed automatically by
{\em comparing} the test samples with {\em ideal} samples using a
 metric, 
\[
\d \DEF
 \left (\D_{ij}, \P_{ij}, \R_{ij} \right )
\]
The metric $\d$ is  suitable  for comparing two spoken words 
 or sentences,
$i$ and $j$.
Note that the metric $\d$ captures both the articulatory($\D_{ij}$) and the
intonation ($\P_{ij}$ and  $\R_{ij}$) capability of the
speaker, both of which together characterise the speaking style of the spoken
speech. 
While $\D_{ij}$ captures the closeness of the content of the two spoken words or
sentences,  
 $\P_{ij}$ and $\R_{ij}$ capture the closeness in terms of {\em intonantion} or the
style of the speaking.
Note that $\P_{ij}$
depends on the parameter {\em  pitch} while the measure $\R_{ij}$ depends on the
{\em stress }
of the spoken speech.

\section{Problem Formulation}

\subsection{Selection of Ideals}

The reference speech samples are analysed qualitatively and assigned a
quantitative measure based on the metric (triplet) discussed in Section
\ref{sec:metric}. Consider
the reference speech data is collected for $\W$ predefined words (or sentences)
from $\G$ classification groups (example, Very Good, Good, Average, Bad, Very Bad) of people. Assume that each group
$\G$ has $N_\G$ number of persons in it. Selection of ideals is based on
initially
segregating the spoken speech samples into 
$\W$ predefined words (or sentences) and a set of $\G$ groups (or
categories) of spoken style categories.
For each  $w \in \W$ and $g \in \G$, determine 
$\overline{\d}_{wg}$, the average over all the utterence by different 
number ($N_\G$) of person. This produces a set of
measurements which represents all the words and groups
namely,
$ \{ \overline{\d_{ij}} \}_{i=1, j=1}^{\W, \G}$ using the pseudo code described
in Algorithm \ref{algo:metric}.

\begin{algorithm}
\begin{algorithmic}
\FOR{i = 1: $\W$} 
\FOR{j = 1: $\G$}
\FOR{k = 1: $N_\G$} 
\FOR{l = 1: $N_\G$} 
\STATE{Calculate $\d^{kl}_{ij}$}
\ENDFOR
\ENDFOR
\STATE{$\overline{\d_{ij}} = \frac{\d^{kl}_{ij}}{(N_\G)^2}$}
\ENDFOR 
\ENDFOR 
\end{algorithmic}
\caption {Computing the metric $\overline{\d_{ij}}$ for the reference speech
data.}
\label{algo:metric}
\end{algorithm}

A reference speech sample $R_{ij}$ is 
chosen  for each word  $i=1, \cdots \W$
and for each group $j = i=1, \cdots \G$ if the {\em variation} within 
each word-group category is {\em not larger} than a predefined threshold. 
Else several (in the worst case
all) reference speech samples in the word-group category are chosen.
The estimation of $\overline{\d_{ij}}$ helps in identifying $R_{ij}$ for each
word-group category.
In essense, $R_{ij}$'s (one or several) are the {\em chosen
representatives} of all the
 speech samples in the $i^{th}$ word and $j^{th}$ group category.
\begin{quote} \em
Note that for each word $i$ and each group $j$ there is a reference speech
sample set
$R_{ij}$ (in the extreme case, this $R_{ij}$ could be multiple files
encompasing
all the reference speech files in that word-group category)  and a score $\overline{\d_{ij}}$.
\end{quote}

\subsection{Classification based on Ideals}
Given a  test speech sample $t$, and $\G$ category groups, the problem is one
of tagging  the given test speech sample ($t$)
to one of the groups $\G$ based on
the closeness of the test speech sample to all the ideals in all the 
$\G$ groups,
 using either the 
$I\!\!L^1$, $I\!\!L^2$ or the $I\!\!L^{\infty}$ metric. In all our experiments
we use $I\!\!L^2$ norm.

It is assumed that the content of the  test speech sample $t$ is known (meaning
that the word or the sentence that has been spoke in known) and is $x$. Now, we
compare the test speech sample with the reference speech samples $R_{xj}$ for
$j = i=1, \cdots, \G$ and calculate the triplet  scores $\T_j = (\D_{tj},
\P_{tj}, \R_{tj})$. Note that  $\T_j$ is obtained by comparing the test sample
$t$ with all the ideals in all the $\G$ groups and then choosing the minimum
$\T_j$.
The test speech sample, $t$, is classified as belonging to the 
group $g$ if the following criteria 
\begin{eqnarray*}
\D_{tg} &\le& \D_{tj}  \\ \nonumber
\P_{tg} &\ge& \P_{tj} \\ \nonumber  
\R_{tg} &\ge& \R_{tj} \nonumber
\label{eq:criteria}
\end{eqnarray*}
is satisfied
$\forall j = 1, \cdots, \G \;\;\; \mbox{and} \;\;\; j \ne g$.

\section{Experimental Results}

A set of $20$ words and $10$ sentences were selected in consultation with
phoneticians and accent training experts. 
The set consisted of words and sentences which were very commonly  prone to
pronunciation error and in some cases the words were  tongue twisters. The
choice of the set is deemed to be capable of assessing the development of
articulation of a person. Data was collected from a set of $20$ people in each
category (very good, good, average, bad and very bad speaking style). All person were asked to speak the
predetermined set of $20$ words and $10$ sentences on the telephone using an
IVR application custom built for collecting data. The speech data was tagged
separately by 
two accent experts into one of the five (very good, good, average,
bad, very bad) categories. 
 Table \ref{tab:two_human} gives the agreement between two human accent
experts.
 Total agreement is when both the human 
experts categorised the same  speech sample as belonging to the same category
(example, both the experts say that the speech sample is good) and 1-step
agreement corresponds to the the human experts differing on their
categorization by a distance of 1 category (example, one  expert say that
the speech sample is good while the other says that the speech sample is
very good or average)

\begin{table}
\begin{center}
\begin{tabular}{|c|c|} \hline
  & Expert 1 - Expert 2 \\ \hline
Total Agreement & 26 \%\\
1-step Agreement & 45 \% \\ \hline
\end{tabular}
\end{center}
\caption{Agreement between two human experts.}
\label{tab:two_human}
\end{table}
For purpose of experimentation to evaluate our system, we divided the speech 
data into $3$ parts.
 We used data 2 parts of the data corresponding
to each of the 5 categories to
select the ideals and used the other 1 part to test the
performance of the system. The overall performance of the system for
classifying spoken speech  is tabulated in Table
\ref{tab:system}.
\begin{table}
\begin{center}
\begin{tabular}{|c||c|c|} \hline
  & Expert 1 - System & Expert 2 - System\\ \hline
Total Agreement &   56 \%& 47 \%\\
1-step Agreement &  100 \% & 90 \%\\ \hline
\end{tabular}
\end{center}
\caption{Agreement between the system and the two human experts.}
\label{tab:system}
\end{table}
The performance of the automated system is {\em
much better} than the performance between two human experts.
Notice that the performance of the human expert - system (see Table
\ref{tab:system}) is better than 
the expert-expert (see Table \ref{tab:two_human}) performance.

\section{Conclusions}
With increase in \ccs\ there is a need for automatic speaking style
analyser. Speaking style analysis by human experts is bound to be biased by
cues that might not necessarily be associated with the speaking style and the
judgement of the speaking style is dependent on the human expert. 
To over come this bias that may be associated with human expert in 
analysing a person for his speaking style we have developed 
a system to automatically analyse the speaking 
style of a person.
We proposed a metric 
which captures both the articulatory capability  and the intonation of the 
speaker, both of which jointly
characterise the speaking style of the person. Experiemntal results show that
the performance of the system far exceeds the performance between two
independent human experts.


\end{document}